\documentclass{article}

\usepackage{PRIMEarxiv}

\usepackage[utf8]{inputenc} 
\usepackage[T1]{fontenc}    
\usepackage{hyperref}       
\hypersetup{hidelinks}

\usepackage{url}            
\usepackage{booktabs}       
\usepackage{amsfonts}       
\usepackage{nicefrac}       
\usepackage{microtype}      
\usepackage{lipsum}
\usepackage{fancyhdr}       
\usepackage{graphicx}       
\graphicspath{{media/}}     

\usepackage{calc}
\usepackage{indentfirst}
\usepackage{epstopdf}
\usepackage{lastpage}
\usepackage{ifthen}
\usepackage{lineno}
\usepackage{float}
\usepackage{amsmath}
\usepackage{setspace}
\usepackage{enumitem}
\usepackage{mathpazo}
\usepackage{booktabs} 
\usepackage[largestsep]{titlesec}
\usepackage{etoolbox} 
\usepackage{tabto} 
\usepackage[table]{xcolor} 
\usepackage{amsthm}
\usepackage{cite}

 \newcounter{problem}
 \newtheorem{Problem}[problem]{Problem}

\usepackage{authblk}
\title{Geometric Learning of Hidden Markov Models via a Method of Moments Algorithm
}
\author[1]{\textbf{Berlin Chen}}
\author[2,\;3]{\textbf{Cyrus Mostajeran}}
\author[4]{\textbf{Salem Said}}
\affil[1\;]{Princeton Neuroscience Institute, Princeton University, USA}
\affil[2\;]{School of Physical and Mathematical Sciences, Nanyang Technological University (NTU), Singapore}
\affil[3\;]{Department of Engineering, University of Cambridge, United Kingdom}
\affil[4\;]{CNRS, Laboratoire Jean Kuntzmann, Université Grenoble-Alpes, Grenoble, France}
\date{}                     
\begin{document}
\maketitle

\begin{abstract}
We present a novel algorithm for learning the parameters of hidden Markov models (HMMs) in a geometric setting where the observations take values in Riemannian manifolds. In particular, we elevate a recent second-order method of moments algorithm that incorporates non-consecutive correlations to a more general setting where observations take place in a Riemannian symmetric space of non-positive curvature and the observation likelihoods are Riemannian Gaussians. The resulting algorithm decouples into a Riemannian Gaussian mixture model estimation algorithm followed by a sequence of convex optimization procedures. We demonstrate through examples that the learner can result in significantly improved speed and numerical accuracy compared to existing learners.
\end{abstract}


\section{Introduction}
\label{sec:Introduction}

\emph{Hidden Markov models} (HMMs) describe states with Markovian dynamics that are hidden in the sense that they are only accessible via observations by a noisy sensor. Specifically, at every time-step $k$, an observation $y_k$ is sampled from an observation space $\mathcal{Y}$ according to the HMM's \emph{observation likelihoods}, which specify the probability of making a particular observation, conditioned on the system being in a certain state. Despite their structural simplicity, HMMs have become a standard tool in the modeling of stochastic time-series \cite{Krishnamurthy2016} in recent decades and have found applications in a wide range of fields including computational biology \cite{Durbin1998,Vidyasagar2014}, signal and image analysis \cite{Cappe2005}, speech recognition \cite{Rabiner1989,Gales2008}, and financial modeling \cite{Mamon2007}.

In order to apply an HMM, it is often necessary to estimate its parameters from data. The standard approach to estimating the parameters of an HMM is using a \emph{maximum likelihood} (ML) criterion. Numerical algorithms for computing the ML estimate are dominated by iterative local-search procedures that aim to maximize the likelihood of observed data, such as the \emph{expectation-maximization} (EM) algorithm \cite{Krishnamurthy2016,Cappe2005}. Unfortunately, these schemes are only guaranteed to converge to local stationary points of the typically non-convex likelihood function and as a result often become trapped in local optima. Thus, to have a chance of converging to a global optimum, a good initialization is usually required. Another drawback of such methods is the significant computational cost associated with long runtimes  due to costly iterations for large datasets.

In order to overcome such challenges, methods of moments have been introduced for HMMs \cite{Chang1996,Mossel2006,Hsu2012,Anandkumar2012,Kontorovich2013,Mattila2017,Huang2018}. Originally, these methods relied on empirical estimation of correlations between consecutive pair- or triplet-wise observations to compute estimates of the HMM parameters. Although computationally attractive, such methods suffered from a loss of accuracy due to a focus on low order correlations in the data. In response, Mattila et al. 
\cite{Mattila2020,Mattila2020thesis} extended these methods to include non-consecutive correlations in the data, resulting in improved accuracy while retaining their attractive computational properties.

\subsection{Hidden Markov models with manifold-valued observations}
\label{HMMs manifold-valued Intro}

The development and analysis of statistical procedures and optimization algorithms on manifolds and nonlinear spaces more broadly have been the subject of intense and growing research interest in recent decades due to the ubiquity of manifold-valued data in a wide range of applications \cite{Absil2009,Barachant2012,Manopt2014,Pennec2020,Geomstats2020,Mostajeran2020,VanGoffrier2021}. Since the application of Euclidean algorithms to such data often has a significantly negative impact on the accuracy and interpretability of the results, it is necessary to devise algorithms that respect the intrinsic geometry of the data. In this work, we turn our attention to HMMs with observations in a Riemannian manifold \cite{SaidHMM2021,Tupker2021}. In particular, we restrict our attention to the class of models with observations in Riemannian symmetric spaces of non-positive curvature, which include hyperbolic spaces, as well as spaces of real, complex, and quaternionic positive definite matrices. We have three motivations for this restriction: (1) standard operations on such spaces have relatively favorable computational properties due to symmetries, (2) there exists a theory of Riemannian Gaussian distributions on such spaces together with associated algorithms such as Riemannian Gaussian mixture estimation \cite{Said2017,Said2018}, and (3) they are applicable to a substantial class of problems involving manifold-valued data, including applications with data in the form of covariance matrices \cite{Said2018}. 

\subsection{Contributions and paper outline}

Our main contribution in this paper is to extend the second-order method of moments algorithm with non-consecutive correlations developed by Mattila et al. \cite{Mattila2020,Mattila2020thesis} to the setting of HMMs with observations in a Riemannian symmetric space of non-positive curvature, where the observation likelihoods take the form of Riemannian Gaussians \cite{Said2018,HOS2022}. The paper is organized as follows. In Section \ref{sec:HMM on manifolds}, we describe HMMs with manifold-valued observations and review the necessary geometric background. In Section \ref{sec:Geometric method of moments}, we review the method of moments algorithms for HMMs and describe how they manifest in the geometric setting. In Section \ref{sec:simulations}, we present a number of simulations based on these algorithms and conclude with a discussion in Section \ref{sec:conclusion}.

\subsection{Notation}

We denote the $i$-th entry of a vector by $[\cdot]_i$, and the element at row $i$ and column $j$ of a matrix by $[\cdot]_{ij}$. Vectors are assumed to be column vectors unless transposed. The vector of all ones is denoted $\mathbf{1}$. We interpret inequalities between vectors and matrices to hold elementwise. The operator $\operatorname{diag}$ acts on vectors and returns the matrix where the vector has been placed on the diagonal, and all other elements set to zero. The matrix Frobenius norm is denoted $\|\cdot\|_F$. The probability of an event $A$ is denoted $\mathbb{P}(A)$.

\section{Hidden Markov models on manifolds}
\label{sec:HMM on manifolds}

We consider a discrete-time hidden Markov model with a finite-state Markov chain on the state space $\mathcal{X}=\{1,\dotsc,N\}$ with time-homogeneous $N\times N$ transition probability matrix $P$ with elements
\begin{equation}
    [P]_{ij}=\mathbb{P}[x_{k+1}=j|x_k=i].
\end{equation}
The initial and stationary distributions of the HMM exist under appropriate assumptions and are denoted by $\pi_0\in\mathbb{R}^N$ and $\pi_{\infty}\in\mathbb{R}^N$, respectively. The HMM is said to be \emph{stationary} if $\pi_0=\pi_{\infty}$. 

We assume that the states are hidden and can only be accessed through observations in a Riemannian symmetric space of non-positive curvature so that the Riemannian Gaussian distribution with probability density function
\begin{equation} \label{Riemannian density}
    p(y|\bar{y},\sigma)=\frac{1}{Z(\sigma)}\exp\left[-\frac{d^2(y,\bar{y})}{2\sigma^2}\right]
\end{equation}
with respect to the Riemannian volume measure $dv(y)$ on $\mathcal{Y}$ is well-defined for any $\bar{y}\in\mathcal{Y}$ and $\sigma>0$, as outlined in \cite{Said2018}. $d(\cdot,\cdot)$ denotes the Riemannian distance function on $\mathcal{Y}$ and $Z(\sigma)$ denotes the normalization factor of the Riemannian Gaussian, whose efficient computation has been the subject of interest in recent years \cite{Heuveline2021,Santilli2021,HOS2022,SalemCyrus2022}. We assume that the observations are sampled from $\mathcal{Y}$ according to conditional probability densities 
\begin{equation}
    B(y_k=y|x_k=j) = p(y|\bar{y}_j,\sigma_j),
\end{equation}
for $j=1,\dotsc,N$ where $p(\cdot|\bar{y}_j,\sigma_j)$ is a Riemannian Gaussian density function of the form (\ref{Riemannian density}) with mean $\bar{y}_j\in\mathcal{Y}$ and dispersion $\sigma_j>0$. 

To use an HMM for applications such as filtering or prediction, its model parameters must be specified or estimated in advance. This task can be formulated as the following learning problem for HMMs:

\begin{Problem} \label{HMM learning problem}
Given a sequence $y_1,\dotsc,y_D$ of observations in 
$\mathcal{Y}$ generated by an HMM of known state space $\mathcal{X}=\{1,\dots,N\}$, estimate the conditional probability densities $B$ and the matrix of transition probabilities $P$.
\end{Problem}

The learning problem is well-posed under the standard assumptions that the HMM is ergodic (irreducible and aperiodic) and identifiable \cite{Cappe2005,Hsu2012,Mattila2020,Mattila2020thesis}. A special case of the learning problem that is worth noting is that of the \emph{known-sensor HMM}, in which the observation likelihoods $B$ are assumed to be known. Known-sensor HMMs are motivated by applications in which the sensor is designed by the user, such as a target tracking system whose sensor specifications can be determined prior to deployment.

Various methods since the inception of HMMs have focused on maximizing the likelihood in terms of both $B$ and $P$; however, recent efforts have demonstrated the potential of methods that decouple the problem \cite{Kontorovich2013,Mattila2017} and estimate $B$ and $P$ sequentially. Specifically, in \emph{parametric-output HMMs} (e.g., Gaussian HMMs), the observation likelihoods are estimated via a general mixture model learner as a first step, followed by identification of the transition matrix $P$ as a second step \cite{Kontorovich2013}. In the first step, assuming that the underlying Markov chain behaves well (e.g. is recurrent) and mixes rapidly, in stationarity, each observation $y_k$ from the HMM can be interpreted as having been sampled from the mixture distribution density
\begin{equation}\label{mixture model}
    p(y)=\sum_{i=1}^N[\pi_{\infty}]_iB(y|\bar{y}_i,\sigma_i).
\end{equation}
Since we are assuming that the observation likelihoods belong to the family of isotropic Riemannian Gaussians on $\mathcal{Y}$, the density (\ref{mixture model}) can be estimated using one of several algorithms for the estimation of mixtures of Riemannian Gaussian distributions including expectation-maximization (EM) \cite{Said2017,Said2018}, stochastic EM \cite{StochasticEM2017}, and online variants \cite{Zanini2017}. The second step is then equivalent to the identification of a known-sensor HMM.

\section{Method of moments algorithms for geometric learning of hidden Markov models}
\label{sec:Geometric method of moments}

\subsection{Method of moments for HMMs}
\label{subsec:Method of moments}

We begin with a brief review of the method of moments algorithm for HMMs developed by Mattila et al. in \cite{Mattila2020}. The significance of this work is that it extends previous method of moments algorithms for HMMs that were based on correlations between consecutive pair- or triplet-wise observations to include non-consecutive correlations in the data. In doing so, the authors improve the accuracy of the approach by reducing the volume of neglected information inherent in the data while maintaining the computationally attractive properties of previous method of moments algorithms.

Before presenting the algorithm in the setting of HMMs with manifold-valued observations, we briefly review a summary of the key steps involved in the second-order algorithm of Mattila et al. \cite{Mattila2020} in the simplest setting where the observations take place in a finite observation alphabet $\{1,\dotsc,Y\}$ with a known $N\times Y$ observation matrix $B$:
\begin{equation} \label{discrete observation matrix}
    [B]_{ij}=\mathbb{P}[y_k=j|x_k=i].
\end{equation}
Methods of moments for HMMs (e.g. \cite{Chang1996,Mossel2006,Hsu2012,Anandkumar2012,Kontorovich2013,Mattila2017,Huang2018}) involve the empirical estimation of low-order correlations in the data, such as pairs $\mathbb{P}[y_k,y_{k+1}]$ or triplets $\mathbb{P}[y_k,y_{k+1},y_{k+2}]$, followed by computation of the HMM parameter estimates by minimizing the discrepancy between the empirical estimates and their analytical expressions via a series of convex optimization problems. In Mattila et al. \cite{Mattila2020}, the authors extend such methods to include non-consecutive correlations of the form $\mathbb{P}[y_k,y_{k+\tau}]$ with $\tau=1,2,\dotsc,\bar{\tau}$ where the number $\bar{\tau}$ is a user-defined lag parameter.

The \emph{lag}-$\tau$ \emph{second-order moments} $M_2(k,\tau)\in\mathbb{R}^{Y\times Y}$ of the HMM are defined as the matrices
\begin{equation}
    [M_2(k,\tau)]_{ij}=\mathbb{P}[y_k=i,y_{k+\tau}=j],
\end{equation}
where $i,j=1,\dotsc,Y$ and $\tau\geq 0$. The case $\tau=0$ reduces to the first-order moments $[M_1(k)]_i=\mathbb{P}[y_k=i]$, where $M_1(k)\in\mathbb{R}^Y$, which for notational convenience is expressed as a special case of second-order moments by writing $M_2(k,0)=\operatorname{diag}(M_1(k))$. For a stationary HMM (i.e., $\pi_0=\pi_{\infty}$) the
lag-$\tau$ second-order moments are related to the HMM parameters according to the equations
\begin{equation} \label{analytic equations}
    M_2(k,\tau)=B^T\operatorname{diag}(\pi_{\infty})P^{\tau}B, \qquad M_2(k,0)=\operatorname{diag}(B^T\pi_{\infty}),
\end{equation}
for any $\tau > 0$ \cite{Mattila2020}.

The lag-$\tau$ second-order moments can be empirically estimated from data as $\hat{M}_2(\tau)$ according to the equation
\begin{equation} \label{M empirical estimation}
    [\hat{M}_2(\tau)]_{ij}=\frac{1}{D-\tau}\sum_{k=1}^{D-\tau}I\{y_k=i,y_{k+\tau}=j\},
\end{equation}
for $\tau = 0,1,\dotsc, \bar{\tau}$, where $D$ is the number of observations and $I$ denotes the indicator function. The next step in the method is moment matching through the minimization of the discrepancy between the empirical estimate $\hat{M}(\tau)$ and its analytical expression by solving the following convex (quadratic) optimization problems:

\begin{enumerate}
    \item Solve 
\begin{equation} \label{moment matching equation 1}
\begin{aligned}
\min_{\hat{\pi}_{\infty}\in\mathbb{R}^{N\times N}} \quad & \|\hat{M}_2(0)-\operatorname{diag}(B^T\hat{\pi}_{\infty})\|_F^2\\
\textrm{s.t.} \quad & \hat{\pi}_{\infty}\geq 0, \quad
\mathbf{1}^T\hat{\pi}_{\infty}=1,\\
\end{aligned}
\end{equation} 
and set $\hat{A}(0)=\operatorname{diag}(\hat{\pi}_{\infty})$. \\
\item For $\tau = 1,\dotsc,\bar{\tau}$, solve 
\begin{equation} \label{moment matching equation 2}
\begin{aligned}
\min_{\hat{P}(\tau)\in\mathbb{R}^{N\times N}} \quad & \|\hat{M}_2(\tau)-B^T\hat{A}(\tau-1)\hat{P}(\tau)B\|_F^2\\
\textrm{s.t.} \quad & \hat{P}(\tau)\geq 0, \quad
\hat{P}(\tau)\mathbf{1}=\mathbf{1},\\
\end{aligned}
\end{equation}
and set $\hat{A}(\tau)=\hat{A}(\tau-1)\hat{P}(\tau)$.
\end{enumerate}
The output of the above moment matching procedure is a sequence $\hat{A}(0),\dotsc,\hat{A}(\bar{\tau})$. In the final step, we use this sequence to estimate the transition matrix $P$ by solving the following least-squares problem, which incorporates information from every lag by construction.
\begin{equation} \label{P estimation}
\begin{aligned}
\min_{\hat{P}\in\mathbb{R}^{N\times N}} \quad & 
\left\lVert
\begin{bmatrix}
\hat{A}(0) \\
\vdots \\
\hat{A}(\bar{\tau}-1)
\end{bmatrix}
\hat{P}
-
\begin{bmatrix}
\hat{A}(1) \\
\vdots \\
\hat{A}(\bar{\tau})
\end{bmatrix}
\right\rVert_F^2\\
\textrm{s.t.} \quad & \hat{P}\geq 0, \quad
\hat{P}\mathbf{1}=\mathbf{1}.\\
\end{aligned}
\end{equation}

The dominant contribution to the computational cost of the above algorithm is independent of the data size $D$ and scales linearly with the number of lags $\bar{\tau}$ included. In contrast, each iteration of the EM algorithm has a complexity of $\mathcal{O}(N^2D)$. In addition to favorable computational properties, it is shown in \cite{Mattila2020,Mattila2020thesis} that the above algorithm is strongly consistent under reasonable assumptions. That is, as the number of samples grows, we expect the estimate of the transition matrix $P$ to converge to its true value.

\subsection{Geometric learning of HMMs using method of moments}
\label{subsec:Geometric learning}

We now return to the problem of estimating the parameters of an HMM with observations in a Riemannian manifold $\mathcal{Y}$ via an extension of the second-order method of moments presented earlier. We assume conditional probability densities to be given by Riemannian Gaussians of the form (\ref{Riemannian density}). The first stage of the process is to estimate the means and variances of the observation densities from data by employing a Riemannian Gaussian mixture learner \cite{Said2018,StochasticEM2017,Zanini2017}. In the case of a known-sensor HMM, this would be unnecessary as the observation densities are known a priori. In the next stage, we use a kernel trick outlined in \cite{Kontorovich2013,Mattila2020thesis} to extend the pairwise correlations between discrete-valued observations $M_2(\tau)$ to an analogous quantity $H(\tau) \in \mathbb{R}^{N\times N}$ applicable in the setting of continuous observation spaces. $H$ is then related to the parameters of the HMM according to the equations
\begin{align}
    H(0) &= \operatorname{diag}(K\pi_{\infty}), \nonumber \\
    H(\tau) &= K^T\operatorname{diag}(\pi_{\infty}) P^{\tau} K, \label{lagged H}
\end{align}
for $\tau = 1, \dotsc, \bar\tau\in\mathbb{N}$, where $\pi_{\infty}$ is the HMM stationary distribution which can be estimated from (\ref{mixture model}), and $K \in \mathbb{R}^{N\times N}$ is defined as 
\begin{equation}\label{K}
    [K]_{ij} = \int_{\mathcal{Y}}B(y\mid x = i)B(y\mid x = j)\,dv(y).
\end{equation} 
The $N\times N$ matrix $K$ in (\ref{K}) is called the the \emph{effective observation matrix} in \cite{Kontorovich2013,Mattila2020thesis} and replaces the $N\times Y$ observation matrix (\ref{discrete observation matrix}). We can compute $K$ using Monte Carlo techniques based on sampling from Riemannian Gaussians \cite{Said2018}.

The elements of the left-hand side of (\ref{lagged H}) can be interpreted as conditional expectations with respect to the joint probability distribution of $y_k$ and $y_{k+\tau}$, which can be empirically estimated from HMM observations as
\begin{align}\label{H estimates}
       [\hat{H}(0)]_{ii} &= \frac{1}{D}\sum_{k=1}^D B(y_k|x=i), \\
    [\hat{H}(\tau)]_{ij} &= \frac{1}{D-\tau}\sum_{k=1}^{D-\tau}B(y_k|x=i)B(y_{k+\tau}|x=j)
\end{align}
in analogy with empirical estimate (\ref{M empirical estimation}) employed in the case of HMMs with a discrete observation space.

Following the estimation of $H(\tau)$ and the computation of $K$, the moment matching procedure now takes the form of minimizing the discrepancy between the empirical estimate $\hat{H}(\tau)$ and the corresponding analytical expressions in (\ref{lagged H}). Specifically, in the case of the known-sensor HMM, we solve the following sequence of convex (quadratic) optimization problems:
\begin{enumerate}
    \item Solve 
\begin{equation} \label{continuous moment matching equation 1}
\begin{aligned}
\min_{\hat{\pi}_{\infty}\in\mathbb{R}^{N\times N}} \quad & \|\hat{H}(0)-\operatorname{diag}(K^T\hat{\pi}_{\infty})\|_F^2\\
\textrm{s.t.} \quad & \hat{\pi}_{\infty}\geq 0, \quad
\mathbf{1}^T\hat{\pi}_{\infty}=1,\\
\end{aligned}
\end{equation} 
and set $\hat{A}(0)=\operatorname{diag}(\hat{\pi}_{\infty})$. \\
\item For $\tau = 1,\dotsc,\bar{\tau}$, solve 
\begin{equation} \label{continuous moment matching equation 2}
\begin{aligned}
\min_{\hat{P}(\tau)\in\mathbb{R}^{N\times N}} \quad & \|\hat{H}(\tau)-K^T\hat{A}(\tau-1)\hat{P}(\tau)K\|_F^2\\
\textrm{s.t.} \quad & \hat{P}(\tau)\geq 0, \quad
\hat{P}(\tau)\mathbf{1}=\mathbf{1},\\
\end{aligned}
\end{equation}
and set $\hat{A}(\tau)=\hat{A}(\tau-1)\hat{P}(\tau)$.
\end{enumerate}
The output is once again a sequence $\hat{A}(0),\dotsc,\hat{A}(\bar{\tau})$, which is used to compute an estimate for the transition matrix $P$ by solving (\ref{P estimation}).

To summarize, the algorithm follows a 2-stage procedure to learn the parameters of an HMM with observations in a Riemannian manifold admitting well-defined Gaussian densities of the form (\ref{Riemannian density}) from data. In stage 1, Riemannian Gaussian mixture estimation is employed to compute estimates for the conditional likelihoods $B$, which are then used in stage 2 to compute an estimate for the transition probabilities $P$ by solving a series of convex optimization problems.
\section{Simulations}
\label{sec:simulations}

We now present the results of several numerical experiments on learning HMMs with manifold-valued observations. In the first example, observations take place in the Poincaré disk model of hyperbolic 2-space. Poincaré models of hyperbolic spaces have been a subject of increasing interest in machine learning in recent years due to their ability to efficiently represent hierarchical data \cite{Nickel2017}. In the second example, we consider a model with observations in the manifold of $2\times 2$ symmetric positive definite (SPD) matrices equipped with the standard affine-invariant Rao-Fisher metric \cite{Said2017}.

\subsection{Example 1: Observations in hyperbolic space}
\label{subsec:Example 1}

We consider the example of an HMM with $N=3$ hidden states with initial distribution $\pi_0=(1,0,0)^T$ and transition matrix
\begin{equation}
    P=\begin{bmatrix}
    0.4 & 0.3 & 0.3 \\
    0.2 & 0.6 & 0.2 \\
    0.1 & 0.1 & 0.8
    \end{bmatrix}
\end{equation}
and observations generated from a Riemannian Gaussian model in the Poincaré disk $\mathcal{Y}=\{y\in\mathbb{C}:|y|<1\}$ with associated means $\bar{y}_1=0$, $\bar{y}_2=0.29+0.82i$, $\bar{y}_3=-0.29+0.82i$ and standard deviations $\sigma_1=0.1$, $\sigma_2=0.4$, $\sigma_3=0.4$ as studied in \cite{SaidHMM2021} in the context of estimation using the EM algorithm. The Riemannian distance function $d(\cdot,\cdot)$ and the Riemannian Gaussian normalization factor $Z(\sigma)$ are given by
\begin{align} 
  d(y, z) = \text{acosh} \left( 1 + \frac{2 |y - z|^2}{(1 - |y|^2)(1 - |z|^2)} \right),  \qquad
  Z(\sigma) = 2\pi\sqrt{\frac{\pi}{2}} \sigma e^{\frac{\sigma^2}{2}} \text{erf} \left( \frac{\sigma}{\sqrt{2}} \right), 
\end{align}
respectively, where $\operatorname{erf}$ denotes the error function \cite{Said2014}.

We employed the second-order method of moments algorithm of Section \ref{subsec:Geometric learning} to learn the parameters of this HMM from observations alone. The model was fitted on 20 HMM chains, each with 10,000 observations. In our implementation, we used the mixture estimation algorithm of \cite{Said2017} to estimate the density (\ref{mixture model}). The full results are reported in Table \ref{EM_compare}, where the true and estimated Gaussian means are denoted by $\bar{y}_i$ and $\hat{y}_i$, respectively. On repeating the experiment with varying $\bar\tau$ and the same random seed—and hence the same estimates for means and dispersions by construction—we observed that incorporating non-consecutive data (i.e., $\bar{\tau} > 1$) up to $\bar{\tau}=3$ significantly improved our estimate for $P$ and produced a more accurate estimate than alternative algorithms \cite{SaidHMM2021,Tupker2021}. Comparing the empirical performance of our algorithm to the numerical results reported in \cite{SaidHMM2021}, we observed that our algorithm performed competitively, while requiring only a fraction of the runtime with the same number of observations. In comparison to the online learning algorithm of \cite{Tupker2021}, which we employed on the same learning problem, we observed improved performance for $\bar{\tau}>1$, with the method of moments algorithm with $\bar{\tau}=3$ producing the most accurate estimate of $P$ out of all considered methods. Interestingly, the runtime of our algorithm was not noticeably affected by the choice of $\bar{\tau}$ in this example since the mixture estimation and computation of $K$ (\ref{K}) accounted for the dominant contribution to the computational cost.

\begin{table}[ht]
\renewcommand{\arraystretch}{1.5}
\centering
\caption{\textit{Comparison of the performance of the method of moments algorithm proposed in this paper against previously published algorithms for estimating HMMs with observations in the Poincaré disk. 
}
\label{EM_compare}}
\begin{tabular}[t]{l>{\centering}p{0.12\linewidth}>{\centering\arraybackslash}p{0.16\linewidth}>{\centering\arraybackslash}p{0.29\linewidth}}
\toprule
 & \textbf{EM algorithm from \cite{SaidHMM2021}} & \textbf{Online algorithm from \cite{Tupker2021}} & 
\textbf{Our proposed algorithm with  (a) $\bar\tau=1$, (b) $\bar\tau=2$, (c) $\bar\tau=3$} \\
\midrule
Mean error, $\left(\sum_id^2(\bar{y}_i, \hat y_i)\right)^{1/2}$ & 0.88 & 0.97 & 0.69\\
Dispersion error, $\left(\sum_i(\sigma_i - \hat \sigma_i)^2\right)^{1/2}$ & 0.42 & 0.37 & 0.34  \\
Transition matrix error, $\| P- \hat{P}\|_F$  & 0.35  & 0.30 & (a) 0.42, (b) 0.26, (c) 0.21 \\
Average runtime & $\sim 1$ hour & $\sim 190$ sec & $\sim 20$ sec \\
\bottomrule
\end{tabular}
\end{table}

\subsection{Example 2: Observations in the manifold of $2\times 2$ SPD matrices with $N=5$ hidden states}
\label{subsec:Example 2}

We now consider an HMM with $N=5$ hidden states that are accessible through noisy observations in the manifold of $2\times 2$ SPD matrices generated from a Riemannian Gaussian model with means $\bar{y}_i$ and standard deviations $\sigma_i$ given in Table \ref{2x2 SPD results}. Here the Riemannian distance function $d(\cdot,\cdot)$ and the Riemannian Gaussian normalization factor $Z(\sigma)$ are given by
\begin{align} \label{SPD distance and Z}
  d(y,z)= \|\log(y^{-1/2}zy^{-1/2})\|_F,  \qquad
  Z(\sigma) = (2\pi)^{\frac{3}{2}}\sigma^2 e^{\frac{\sigma^2}{4}} \text{erf} \left( \frac{\sigma}{2} \right). 
\end{align}
While the expression for the Riemannian distance function holds true for higher dimensional SPD matrices, the analytical expression for $Z(\sigma)$ in (\ref{SPD distance and Z}) is only valid in the $2\times 2$ case. Nonetheless, $Z(\sigma)$ can be directly computed or approximated for higher dimensional SPD matrices \cite{Said2017,Said2018,Heuveline2021,Santilli2021,HOS2022,SalemCyrus2022}.

\begin{figure}[htb!]
\begin{center}
\includegraphics[width=\textwidth]{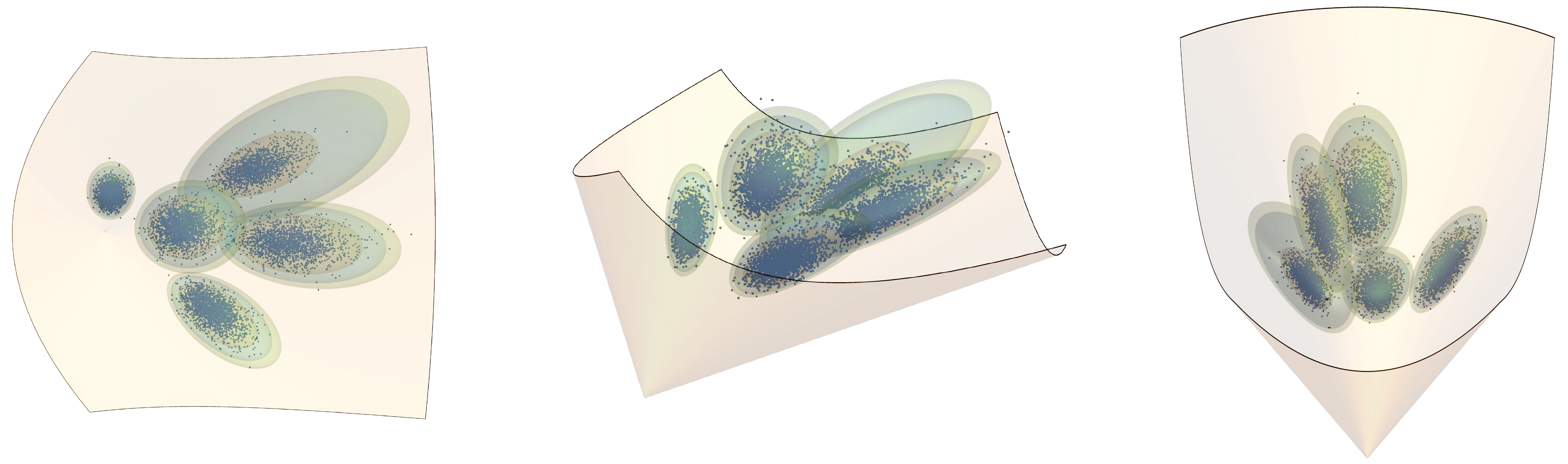}
\end{center}
\caption{Visual representation of the Riemannian Gaussian model estimated from 10,000 observations from three vantage points: top view (left), side view (middle), and front view (right). Each $2\times 2$ SPD-valued observation is plotted as a point in the interior of the pointed convex cone $\{(a,b,c)\in\mathbb{R}^3: a\geq 0,\, ac-b^2\geq 0\}$. The shaded compact regions within the cone are superlevel sets of the 5 estimated Riemannian Gaussian densities that represent the observation likelihoods.}
\label{SPD figure}
\end{figure}

The transition matrix $P$ of the underlying Markov chain is
\begin{equation}
P=
    \begin{bmatrix}
    0.3 & 0.1 & 0.2 & 0.1 & 0.3 \\
    0.1 & 0.4 & 0.2 & 0.2 & 0.1 \\
    0.2 & 0.2 & 0.3 & 0.1 & 0.2 \\
    0.1 & 0.1 & 0.2 & 0.5 & 0.1 \\
    0.4 & 0.1 & 0.1 & 0.1 & 0.3
    \end{bmatrix}.
\end{equation}
We employed our proposed geometric second-order method of moments algorithm with $\bar{\tau}=1$ to sequentially estimate the underlying Gaussian model and the probability transition matrix from 10,000 observations. The results of the Gaussian mixture estimation procedure are reported in Table \ref{2x2 SPD results} and demonstrate a high level of accuracy. The estimated Riemannian Gaussian model with means $\hat{y}_i$ and standard deviations $\hat{\sigma}_i$ as well as the observations used to learn the model are visualized in Figure \ref{SPD figure}.

\begin{table}[ht]
\renewcommand{\arraystretch}{1.5}
\centering
\caption{\textit{True and estimated Riemannian Guassian mixture model parameters. $\hat{y}_i$ and $\hat{\sigma}_i$ denote the estimated Riemannian Gaussian means and standard deviations, respectively. $\pi_{\infty}$ and $\hat{\pi}_{\infty}$ denote the true and estimated stationary distributions, respectively.}
\label{2x2 SPD results}}
\begin{tabular}{p{0.03\linewidth}>{\centering\arraybackslash}p{0.15\linewidth}>{\centering\arraybackslash}p{0.15\linewidth}>{\centering\arraybackslash}p{0.18\linewidth}>{\centering\arraybackslash}p{0.15\linewidth}>{\centering\arraybackslash}p{0.18\linewidth}}
\toprule
& $i=1$ & $i=2$ & $i=3$ & $i=4$  & $i=5$  \\
\midrule\\
\addlinespace[-2ex]
$\bar{y}_i$ & 
$\begin{bmatrix}  1.646 & 0.056 \\ 0.056 & 2.379 \end{bmatrix}$ &
$\begin{bmatrix} 2.294 & 0.744 \\ 0.744 & 1.415 \end{bmatrix}$ & 
$\begin{bmatrix}  2.631 & -0.127 \\ -0.127 &  1.277 \end{bmatrix}$ &
$\begin{bmatrix} 0.674 & 0.454 \\ 0.454 & 2.056 \end{bmatrix}$ &
$\begin{bmatrix}  1.829 & -0.919 \\ -0.919 &  1.602 \end{bmatrix}$ \\
\addlinespace[1.5ex]
$\hat{y}_i$ & 
$\begin{bmatrix}  1.642 & 0.051 \\ 0.051 & 2.383 \end{bmatrix}$ &
$\begin{bmatrix} 2.300 & 0.743 \\ 0.743 & 1.412 \end{bmatrix}$ & 
$\begin{bmatrix}  2.642 & -0.128 \\ -0.128 &  1.277 \end{bmatrix}$ &
$\begin{bmatrix} 0.672 & 0.454 \\ 0.454 & 2.057 \end{bmatrix}$ &
$\begin{bmatrix}  1.830 & -0.920 \\ -0.920 &  1.604 \end{bmatrix}$ \\
\addlinespace[1.5ex]
$\sigma_i$ & $0.1$ & $0.1$ & $0.1$ & $0.1$ & $0.1$ \\
\addlinespace[1.5ex]
$\hat{\sigma_i}$ & $0.099$ & $0.100$ & $0.099$ & $0.101$ & $0.101$ \\
\addlinespace[1.5ex]
$\pi_{\infty}$  & $0.227$ & $0.171$ & $0.199$ & $0.195$  & $0.207$ \\
\addlinespace[1.5ex]
$\hat{\pi}_{\infty}$ & $0.229$ & $0.159$ & $0.201$ & $0.195$  & $0.216$ \\
\bottomrule
\end{tabular}
\end{table}

The estimated transition matrix $\hat{P}$ is 
\begin{equation}
\hat{P} =
    \begin{bmatrix}
    0.291 & 0.088 & 0.195 & 0.092 & 0.334 \\
    0.104 & 0.409 & 0.185 & 0.188 & 0.114 \\
    0.199 & 0.206 & 0.297 & 0.098 & 0.200 \\
    0.091 & 0.113 & 0.202 & 0.482 & 0.112 \\
    0.407 & 0.105 & 0.106 & 0.083 & 0.299
    \end{bmatrix},
\end{equation}
which yields a relative approximation error of
\begin{equation}
    \frac{\|P-\hat{P}\|_F}{\|P\|_F} = 0.050
\end{equation}
with respect to the Frobenius norm. The mean error in the estimated transition probabilities is 
\begin{equation}
    \frac{1}{N^2}\sum_{i,j=1}^N|[P]_{ij}-[\hat{P}]_{ij}|\approx 0.01.
\end{equation}


\section{Conclusion}
\label{sec:conclusion}

In this paper, we have shown that the recent method of moments algorithms for HMMs can be generalized to geometric settings in which observations take place in Riemannian manifolds. We observe through simple numerical simulations that the documented advantages of method of moments algorithms, including their competitive accuracy and attractive computational and statistical properties, may continue to hold in the geometric setting. Nonetheless, we expect unique computational challenges to arise in applications involving high-dimensional Riemannian manifolds. Specifically, using Markov chain Monte Carlo (MCMC) algorithms to compute the effective observation matrix $K$ defined in (\ref{K}) may become prohibitively expensive in high dimensions, which is not the case in the Euclidean setting as $K$ admits a closed form analytic expression for multivariate Gaussian HMMs. Thus, a key technical challenge for the effective application of the proposed algorithm in problems involving high-dimensional manifolds is to devise algorithms for the efficient and scalable computation of $K$. Further developments of the approach may include extensions to models that incorporate third- or higher-order moments or more elaborate dynamics and control inputs.

\section*{Acknowledgments}
B.C. acknowledges funding from the Faculty of Mathematics at the University of Cambridge as part of the Cambridge Mathematics Placements (CMP) Programme. C.M. was supported by an NTU Presidential Postdoctoral Fellowship and an Early Career Research Fellowship at Fitzwilliam College, Cambridge.

\bibliographystyle{unsrt}  
\bibliography{IEEEXplore2020}

\end{document}